\documentclass[letterpaper, 10 pt, conference]{ieeeconf}  

\IEEEoverridecommandlockouts                              

\usepackage[T1]{fontenc}
\usepackage{amsmath} 
\usepackage{amssymb}  
\usepackage{amsfonts}
\usepackage{bbm}
\usepackage{dsfont}
\usepackage{graphicx}
\usepackage[font={rm,md,up}]{subfig}
\usepackage{float}
\usepackage{multirow}
\usepackage{url}
\usepackage{arydshln}
\usepackage{xcolor}

\DeclareMathOperator*{\argmin}{arg\,min}

\makeatletter
\newcommand*\bigcdot{\mathpalette\bigcdot@{.5}}
\newcommand*\bigcdot@[2]{\mathbin{\vcenter{\hbox{\scalebox{#2}{$\m@th#1\bullet$}}}}}
\makeatother

\newcommand\blfootnote[1]{%
  \begingroup
  \renewcommand\thefootnote{}\footnote{#1}%
  \addtocounter{footnote}{-1}%
  \endgroup
}

\title{\LARGE \bf
Cross-Subject Statistical Shift Estimation for Generalized Electroencephalography-based Mental Workload Assessment}

\author{Isabela Albuquerque$^{1,*}$, Jo\~ao Monteiro$^{1}$, Olivier Rosanne$^{1}$, Abhishek Tiwari$^{1}$, Jean-Fran\c{c}ois Gagnon$^{2}$, \\ and Tiago H. Falk$^{1}$ \\ 
$^{1}${\it{Institut National de la Recherche Scientifique,
        Universit\'{e} du Qu\'{e}bec, Montreal, Canada}} \\
$^{2}${\it{Thales Research and Technology Canada, Qu\'{e}bec, Qu\'{e}bec, Canada}}\\
$^{*}${\tt\small isabelamcalbuquerque@gmail.com}}

\begin{document}

\maketitle
\thispagestyle{empty}
\pagestyle{empty}

\begin{abstract}
Assessment of mental workload in real-world conditions is key to ensure the performance of workers executing tasks that demand sustained attention. Previous literature has employed electroencephalography (EEG) to this end despite having observed that EEG correlates of mental workload vary across subjects and physical strain, thus making it difficult to devise models capable of simultaneously presenting reliable performance across users. Domain adaptation consists of a set of strategies that aim at allowing for improving machine learning systems performance on unseen data at training time. Such methods, however, might rely on assumptions over the considered data distributions, which typically do not hold for applications of EEG data. Motivated by this observation, in this work we propose a strategy to estimate two types of discrepancies between multiple data distributions, namely marginal and conditional shifts, observed on data collected from different subjects. Besides shedding light on the assumptions that hold for a particular dataset, the estimates of statistical shifts obtained with the proposed approach can be used for investigating other aspects of a machine learning pipeline, such as quantitatively assessing the effectiveness of domain adaptation strategies. In particular, we consider EEG data collected from individuals performing mental tasks while running on a treadmill and pedaling on a stationary bike and explore the effects of different normalization strategies commonly used to mitigate cross-subject variability. We show the effects that different normalization schemes have on statistical shifts and their relationship with the accuracy of mental workload prediction as assessed on unseen participants at training time. 
\end{abstract}

\section{Introduction}

Monitoring \blfootnote{\textbf{This work has been submitted to the IEEE for possible publication. Copyright may be transferred without notice, after which this version may no longer be accessible.}} mental workload in a fast and accurate manner is critical in scenarios where the full attention of an individual is fundamental for the security of others. Firefighters, air traffic controllers, and first responders, for instance, are constantly exposed to such work conditions. In many cases, in addition to demanding mental tasks, individuals are also under varying levels of physical strain. Measuring mental workload under such scenarios is challenging, especially when relying on wearable sensors \cite{albuquerque2018analysis}. 

Passive brain-computer interfaces (BCIs) have been widely used in the past for mental workload monitoring (e.g., \cite{zhang2014recognition, zhang2017nonlinear, wang2015using}). Existing models, however, exhibit high cross-subject variability, hence hindering their applicability in real-world scenarios. As pointed out in \cite{yin2017cross}, models are usually subject-specific and present poor generalization when training and testing conditions are distinct in terms of the represented individuals. Anatomic and environmental factors have been attributed as the main causes of the cross-subject variability \cite{wei2018subject, wu2015online, wu2015reducing}. Additionally, shifts between training and testing conditions could occur due to different data collection equipment, as well as changes in the electrodes positioning during an experimental session or even in the performance of each individual for the same task. 

A standard way of compensating for the high cross-subject variability with EEG-based passive BCIs is to \textit{calibrate} the model prior to applying it to an unseen individual. This is achieved by collecting a (usually small) number of labelled examples from this particular user and retraining or pruning the model to fine-tune it to the new user \cite{lotte2015signal}. Recent work, however, has highlighted that this calibration step can be too costly and time-consuming, hence not very practical \cite{lotte2018review, arico2018passive}. Improving the cross-subject generalization of current BCIs is therefore critical for real-world applications, such as mental workload monitoring. 

An alternative strategy to calibrating BCIs to unseen subjects/conditions is to develop methods that reduce the variability between training and testing conditions. To this end, methods such as domain adaptation (DA) have been proposed \cite{sun2017correlation, daume2006domain}. A standard DA strategy corresponds to augmenting the learning objective of an algorithm with a term that accounts for how \textit{invariant} the current model is with respect to data from different distributions \cite{ganin2016domain, Tzeng_2017_CVPR}. The goal of this regularization term is to enforce the learned model to ignore domain-specific cues. It is important to emphasize that throughout the remainder of this paper, the terms {\emph{domain}} and {\emph{distribution}} will be used interchangeably.

Previous work on domain adaptation has shown that different techniques rely on distinct assumptions over the training and testing distributions \cite{ben2010impossibility, zhao2019learning}. For example, a common requirement is \textit{covariate shift} assumption, which considers that the distributions of labels $y$ conditioned on data $x$, $p(y|x)$, do not shift across training and testing conditions and only the marginal distributions $p(x)$ shift \cite{ben2010impossibility}. In the case of EEG-based passive BCI applications, however, previous work has argued that $p(y|x)$ is likely to shift between different subjects \cite{wu2016online, wu2015reducing, wu2015online, wei2018subject}. Therefore, the covariate shift assumption cannot be taken for granted since, given feature vectors $x_1$ and $x_2$ respectively acquired from two distinct subjects and represented in a shared feature space, $p_1(y|x_1) \neq p_2(y|x_2)$ even in the case where $x_1=x_2$. As discussed in \cite{zhao2019learning, johansson2019support}, when the covariate shift assumption does not hold, there is a trade-off between learning domain-invariant representations and obtaining a small prediction error across different domains that needs to be optimized.

Verifying whether the underlying assumptions of a particular approach hold in practice is a frequently overlooked step by domain adaptation approaches \cite{akuzawa2018domain}. In this work, we claim that it is necessary to evaluate the underlying structure of a particular dataset in order to verify which types of distribution shifts exist and which assumptions could be safely considered (or not), when utilizing domain adaptation strategies. To this end, our main contributions are: (i) We introduce a method to estimate the cross-subject mismatch between the conditional label distributions; (ii) We apply a notion of divergence introduced in \cite{kifer2004detecting} to estimate the mismatch between marginal distributions of pairs of subjects; (iii) We investigate whether common practices in the EEG literature to mitigate cross-subject variability, such as normalizing spectral features, are able to mitigate both conditional and marginal distributional shifts.

Given the relevance of mitigating cross-subject variability on EEG-based mental workload assessment, we empirically validate our proposed method on the WAUC dataset \cite{albuquerque2020wauc}. The dataset is comprised of EEG data collected during a mental workload modulation task with subjects performing different activity levels and activity types. In this contribution, we extend our first efforts towards quantifying cross-subjects statistical shifts as presented in \cite{albuquerque2019cross}, by considering a larger number of subjects in our analysis (total of 18 subjects), and, more importantly, we investigate how different ways to modulate physical activity affect the cross-subject statistical shifts on EEG correlates of mental workload.

The remainder of this paper is organized as follows: in Section \ref{sec:da} we provide an overview of domain adaptation and formalize the problem of generalizing across subjects under this setting. In Section \ref{sec:cs}, the proposed strategies to estimate conditional and marginal shifts are presented. In Sections \ref{sec:exps} and \ref{sec:results}, we describe the experimental setup and present the results, respectively. Finally, we outline the main conclusions in Section \ref{sec:conc}. 

\section{Domain Adaptation and Cross-subject generalization}\label{sec:da}

Consider $d$-dimensional feature vectors $x \in \mathbb{R}^d$, computed from data through a deterministic mapping, such as power spectral density computations from EEG signals. We denote the feature space as $\mathcal{X}$. Further consider a labeling function $f:\mathcal{X}\rightarrow\mathcal{Y}$, where the label space is represented by $\mathcal{Y}$. For example, $\mathcal{Y}$ would be $\{0,1\}$ for a binary classification case. A domain $\mathcal{D}$ is defined as a distribution over $\mathcal{X}$.

Moreover, let a hypothesis $h$ be a mapping $h:\mathcal{X}\rightarrow\mathcal{Y}$, such that $h \in \mathcal{H}$, where $\mathcal{H}$ is a set of candidate hypothesis, or a hypothesis class. Finally, we define the risk $R$ associated with a given hypothesis $h$ on domain $\mathcal{D}$ as:
\begin{equation}
    R[h] = \mathbb{E}_{x \sim \mathcal{D}} \ell [h(x), f(x)],
\end{equation}
where the loss $\ell:\mathcal{Y} \times \mathcal{Y} \rightarrow R_{+}$ quantifies how different $h$ is from the true labeling function $f$ on $\mathcal{D}$. Supervised learning can be defined as searching the minimum risk hypothesis $h^*$ within $\mathcal{H}$, i.e.,:
\begin{equation}
    h^* = \argmin_{h \in \mathcal{H}} R[h].
\end{equation}
However, computing $R[h]$ is generally intractable since one does not usually have access to $\mathcal{D}$, but instead just observed samples from the domain.

\subsection{Empirical risk minimization}
Given the intractability of the risk minimization setting described above, empirical risk minimization is a common practical alternative framework for supervised learning. In such case, a sample $X$ of size $N$ is observed from $\mathcal{D}$, i.e., $X = \{x_1, x_2, \dots, x_N\}$, where all $x_n$ are assumed to be independently sampled from the domain $\mathcal{D}$ (i.e., the i.i.d. assumption holds). The empirical risk is thus defined as:
\begin{equation}
    \hat{R}_X [h_X] = \frac{1}{n}\sum\limits_{i=1}^{n} \ell[h_S(x_i),f(x_i)],
\end{equation}
and the generalization error (or generalization gap) will be the difference between the true and empirical risks, i.e., \mbox{$\epsilon = | R[h_X] - \hat{R}_X [h_X] |$}. Ideally , $\hat{R}_X [h_X] \approx 0$ and $\epsilon \approx 0$, in which case $h_S$ is able to attain a low risk across new samples of $\mathcal{D}$, not observed at training time.

\subsection{Domain adaptation}
We now analyze the case such that the i.i.d. assumption, which considers $x_n$ in $X$ are all sampled according to a fixed domain $\mathcal{D}$, does not hold. More specifically, we assume that a set of $M$ different domains exist. In the following, we describe two recent results and formally define the statistical shifts that might be observed when different domains are considered.

Since most relevant results and theoretical guarantees were proven specifically for the case in which $M=2$, we consider such setting and define two domains, referred as the source and target domains $\mathcal{D}_S$ and $\mathcal{D}_T$, respectively. A bound for the risk of a given hypothesis on the target domain $R_T[h]$ was introduced in \cite{ben2007analysis}. This result shows that $R_T[h]$ depends on $R_S[h]$, the risk of $h$ on the source domain, a notion of divergence between both domains, as well as the minimum risk that can be achieved by some $h \in \mathcal{H}$ on both $\mathcal{D}_S$ and $\mathcal{D}_T$. We restate this result in the following Corollary.
\\
\\
\noindent
\textit{Corollary 1 (Ben-David et al. \cite{ben2007analysis}, Theorem 1):} Consider two domains $\mathcal{D}_S$ and $\mathcal{D}_T$ over a shared feature space. The risk of a given hypothesis $h$ on the target domain will be thus bounded by:
\begin{equation}
    R_T[h] \leq R_S[h] + d_{\mathcal{H}\Delta\mathcal{H}}[\mathcal{D}_S, \mathcal{D}_T] + \lambda,
\end{equation}
where $\lambda$ accounts for how ``adaptable'' the class $\mathcal{H}$ is and it is defined as the minimal total risk over both domains that can be achieved by some $h \in \mathcal{H}$: 
\begin{equation}
    \lambda = \min_{h \in \mathcal{H}}[R_S[h]+R_T[h]].
\end{equation}
The term $d_{\mathcal{H}\Delta\mathcal{H}}[\mathcal{D}_S, \mathcal{D}_T]$ corresponds to the $\mathcal{H}\Delta\mathcal{H}$-divergence introduced in \cite{kifer2004detecting} for a hypothesis class $\mathcal{H}\Delta\mathcal{H}=\{h(x) \oplus h'(x) | h, h' \in \mathcal{H}\}$, where $\oplus$ is the XOR operation.

An extension of that result was introduced in \cite{zhao2019learning} in order to replace $\lambda$ by a term that explicitly accounts for a possible mismatch between the labeling rules of source and target domains, denoted as $f_S$ and $f_T$, respectively. For that, the divergence between source and target is computed over a hypothesis class $\mathcal{\tilde{H}}$ defined as $\mathcal{\tilde{H}} = \{sign(|h(x) - h'(x)| - t) | h, h' \in \mathcal{H},0\leq t\leq1\}$. We state this result in the following Corollary:
\\
\\
\noindent
\textit{Corollary 2 (Zhao et al. \cite{zhao2019learning}, Theorem 4.1):}
\begin{equation}\label{eq:gordon}
\begin{split}
    & R_T[h] \leq R_S[h] + d_{\mathcal{\tilde{H}}}[\mathcal{D}_S, \mathcal{D}_T] +  \\ & \min \{\mathbb{E}_{x \sim \mathcal{D}_S}\mathds{1}[f_S(x) \neq f_T(x)],
    \mathbb{E}_{x \sim \mathcal{D}_T}\mathds{1}[f_S(x) \neq f_T(x)]\},    
\end{split}
\end{equation}
where $\min \{\mathbb{E}_{x \sim \mathcal{D}_S}\mathds{1}[f_S(x) \neq f_T(x)], \mathbb{E}_{x \sim \mathcal{D}_T}\mathds{1}[f_S(x) \neq f_T(x)]\}$ accounts the mismatch between the labeling functions. 

In light of Corollaries 1 and 2, it is possible to point out the two main aspects that determine how well a hypothesis $h$ generalizes from the source to the target domain. For that, the input space $\mathcal{X}$ must be such that the divergence $d_{\mathcal{H}}[\mathcal{D}_S, \mathcal{D}_T]$ between the marginal distributions is low, while the mismatch between labeling functions accounted by the term $\min \{\mathbb{E}_{x \sim \mathcal{D}_S}\mathds{1}[f_S(x) \neq f_T(x)], \mathbb{E}_{x \sim \mathcal{D}_T}\mathds{1}[f_S(x) \neq f_T(x)]\}$ is also small. Previous work on domain adaptation(e.g. \cite{ganin2016domain}) has mostly focused on mitigating the mismatch between marginal distribution and assumed that labeling functions were the same across domains. However, when this is not case, decreasing the discrepancy between marginal distributions \cite{zhao2019learning} or adding more data \cite{crammer2008learning} might actually hurt the performance of a model on the target domain.  

\subsection{Cross-subject generalization as domain adaptation}
In this work, we formalize the problem of learning passive BCIs that generalize across subjects under the domain adaptation setting. For that, consider a dataset with a total of $M$ subjects and that each subject is associated with domain $\mathcal{D}_i$ and labeling function $f_i$, $\forall \, i=\{1, \cdots, M\}$. Without loss of generality, assume that recordings from the first $M-1$ subjects are available at training time and we are interested in predicting how well a hypothesis $h \in \mathcal{H}$ would perform in the $M$-th subject, which was not considered at training time. Let $\mathcal{D}_S = \bigcup_{k=1}^{M-1} \mathcal{D}_k$ be the source domain defined as the union of the domains corresponding to the training subjects. Taking into consideration Equation \ref{eq:gordon}, we can bound the risk on the $M$-th unseen subject, $R_M[h]$ as  

\begin{equation}\label{eq:bound_eeg}
\begin{split}
    & R_M[h] \leq R_S[h] + d_{\mathcal{\tilde{H}}}[\mathcal{D}_S, \mathcal{D}_M] +  \\ & \min \{\mathbb{E}_{x \sim \mathcal{D}_S}\mathds{1}[f_S(x) \neq f_M(x)],
    \mathbb{E}_{x \sim \mathcal{D}_M}\mathds{1}[f_S(x) \neq f_M(x)]\}.    
\end{split}
\end{equation}
In practice, we aggregate the available test samples from all the training subjects to estimate the risk of $h$ in the source domain $R_S[h] = \sum_{k=1}^{M-1}R_i[h]$, i.e. $\$$. However, there is no such straightforward way of estimating the two remaining terms of the bound. In the next Section, a strategy to compute these two terms is proposed.

\section{Estimating shifts across multiple distributions}
\label{sec:cs}
In this Section we provide practical strategies to estimate both conditional and marginal shifts for a case where multiple domains (subjects) are available. Quantifying such mismatch will enable us to:
\begin{itemize}
    \item Shed light on which domain adaptation strategies should be used for a given scenario by verifying whether, for example, the covariate shift assumption holds.
    \item As these quantities are related to how well a particular hypothesis will perform on unseen subjects, we can use their estimates computed considering different feature spaces and infer which one would achieve better performance on unseen subjects.
\end{itemize}

\subsection{Conditional shift}
A conditional shift is observed across subjects when the labeling function (or, in the stochastic case, the conditional distribution of the labels given the input features) differ among the subjects, i.e., for $M$ subjects, we have $f_i(x) \neq f_j(x)$, $\forall i,j=\{1, \cdots, M\}$. In order to characterize the cross-subject conditional shift of a dataset of $M$ subjects, we consider the following quantity on the generalization bound presented in Corollary 2 for all pairs of subjects:
\begin{equation}\label{expecs}
\min \{ \mathbb{E}_{\mathcal{D}_i}[|f_i - f_j |], \mathbb{E}_{\mathcal{D}_j}[|f_j - f_i|] \},
\end{equation}
where $i,j=\{1, \cdots, M\}$. In practice, it is not possible to compute such quantity as one does not have access to the true labeling functions and computing the expectations in Eq. \ref{expecs} is intractable. 

We thus propose to estimate such values by learning a labeling rule for each one of the domains, and account for how well it classifies examples from the other domain. Assuming that we are able to learn a good predictor for the labels of each domain, such approach is capable of accounting for how ``close'' the true labeling functions of different domains are. In practice, we consider that two labeled samples of size $N$ from domains $i$ and $j$ are available and compute the following estimator $\mu_{i,j}$ for the quantity $\mathbb{E}_{\mathcal{D}_i}[|f_{i} - f_{j}|]$:
\begin{equation}
\mu_{i,j} = \frac{1}{N} \sum_{n=1}^{N} \mathds{1}[f_i(x_n^i) \neq \tilde{f}_j(x_n^i)],
\end{equation}
where $(x^i_n, y^i_n) \sim \mathcal{D}_i$, and $\tilde{f}_j$ is an approximated labeling function for the j-th subject. We decided to have as $\tilde{f}_j$ a non-parametric decision procedure based on the Euclidean distance between data points in a fixed feature space. For that, we use a k-nearest neighbor (k-NN) labeling function, i.e., a k-NN binary classifier trained on $\mathcal{D}_j$ to classify as low or high mental workload condition data sampled from $\mathcal{D}_i$. Based on $\mu_{i,j}$ and $\mu_{j,i}$ we estimate the value $d_{i,j} = d_{j,i}$ = $\min\{\mu_{i,j}, \mu_{j,i}\}$ and compose a Hermitian (elements symmetric with respect to the main diagonal are equal) disparity matrix $D$ defined as:

\begin{equation}
D = 
\begin{bmatrix}
    d_{1,1} & d_{1,2} & \dots  & d_{1,M} \\
    d_{2,1} & d_{2,2} & \dots  & d_{2,M} \\
    \vdots  & \vdots & \ddots  & \vdots  \\
    d_{M,1} & d_{M,2} & \dots  & d_{M,M}
\end{bmatrix}.
\end{equation}
Notice that in the case we obtain optimal approximate labeling functions, i.e., $f_i(x_n^i) = \tilde{f}_j(x_n^i)$, $\forall i=j$, the trace of $D$ is equal to 0. Finally, in order to obtain a single value representing the conditional shift of all subjects in a dataset, we aggregate the values of pairwise conditional shifts. For that, we compute the Frobenius norm $||.||_F$ of the disparity matrix $D$:  
\begin{equation}\label{eq:frob}
    ||D||_F = \sqrt{\sum_{i=1}^M\sum_{j=1}^M |d_{i,j}|^2}.
\end{equation}
The resulting $||D||_F$ is then rescaled to the $[0, 1]$ interval to allow for easier comparison across feature spaces. \\

\subsection{Marginal shift}
The $\mathcal{H}$-divergence between two distributions $\mathcal{D}_S$ and $\mathcal{D}_T$ is defined as: 
\begin{equation}
        d_{\mathcal{H}}[\mathcal{D}_S, \mathcal{D}_T] = 2 \sup_{\eta \in \mathcal{H}} | \text{Pr}_{x\sim\mathcal{D}_S} [\eta(x)=1] 
        - \text{Pr}_{x\sim\mathcal{D}_T} [\eta(x)=1] |.
\end{equation}
As discussed in \cite{ben2007analysis}, $d_{\mathcal{H}}[\mathcal{D}_S, \mathcal{D}_T]$ can be estimated from the error $\epsilon$ of a binary classifier trained to distinguish samples from $\mathcal{D}_S$ and $\mathcal{D}_T$. The lower $\epsilon$ is, the highest the estimate of $d_{\mathcal{H}}$ will be, since in this case, there is a hypothesis $\eta$ capable of distinguishing between $\mathcal{D}_S$ and $\mathcal{D}_T$ with high accuracy. Notice that the $\mathcal{H}$-divergence only accounts for discrepancies between the marginal distributions of the domains, not accounting for how each data point is labeled. Therefore, it is not necessary to have access to labeled samples from the considered domains to estimate its value.

Our proposed approach to estimate the cross-subject marginal shift from a group of $M$ domains (subjects) relies on estimating pair-wise domain divergences, i.e., we compute $d_{\mathcal{H}}[\mathcal{D}_i, \mathcal{D}_j]$ $\forall i,j=\{1, \cdots, M\}$. In the case of scenarios where EEG datasets are taken into account, estimating cross-domain marginal shifts consists in obtaining models capable of performing pair-wise discrimination of features extracted from recordings of different subjects. Similarly to the proposed strategy to estimating cross-subject conditional shift values, we introduce a Hermitian matrix $H$ that accounts for marginal shifts between all subjects. Each entry of $H$ corresponds to the average error rate of pair-wise subject classification. In practice, we use 5-fold cross validation to estimate the error rates. An aggregate value of marginal shift can also be obtained via the rescaled Frobenius norm of $H$.    

\section{Experimental setup} \label{sec:exps}
In this section we provide an overview of WAUC dataset, as well as introduce the features, normalization approaches, and the mental workload classification scheme utilized in the experiments. Moreover, we describe the implementation details in order to allow reproducibility of our experiments.

\subsection{WAUC dataset}
We consider the EEG recordings of the Workload Assessment Under physical aCtivity (WAUC) dataset \cite{albuquerque2020wauc} for our experiments. This dataset was collected when subjects had cognitive and physical workload simultaneously modulated. Mental workload was modulated via the MATB-II task while physical activity consisted of running on a treadmill at 5km$\slash$h or pedalling on a stationary bike at 70rpm. EEG data was recorded using a Neurolectrics Enobio 8-channel wearable headset with a sampling rate of 500Hz. Electrodes were placed following the 10-20 system at the frontal area in the positions AF7, FP1, FP2, and AF8. References were placed at FPz and Nz. The WAUC dataset also contains recordings from baseline periods during the data collection. There are two different types of baseline recordings: 1) EEG was recorded when no mental or physical effort was demanded from the participant (eyes-closed, no movement), and 2) Data was acquired when only physical effort was taken into account, i.e., subjects were running on the treadmill or pedalling at the specified speed while executing no mental task. Subjects performed two experimental sessions, each with an approximate duration of 10 minutes and under a different mental workload level. For our experiments, we considered a total of 18 subjects from the dataset, whom half performed physical activity with the treadmill and the other half with the bike.

\subsection{Feature extraction}
Our preprocessing and feature extraction pipeline consisted in downsampling the EEG recording to 250Hz, filtering it with a band-pass filter from 0.5-45 Hz, and computing features over 4-second epochs with 3 seconds of overlap between consecutive windows. Considering a 10-minute experimental session, after downsampling and epoching the data, we obtained an approximate total of 600 points per subject$\bigcdot$session. As the literature has shown that increases in mental workload incur in changes in alpha, beta, and theta bands in the frontal cortex \cite{pilot, mwl7}, we considered power spectral density (PSD) features in standard EEG frequency bands, namely: delta (0.1-4 Hz), theta (4-8 Hz), alpha (8-12 Hz), and beta (2-30 Hz). 

\subsection{Normalization}
Feature normalization is a common practice used to minimize the effects of cross-subject variability for EEG-based classification tasks. Task-based Features are typically normalized with respect to the statistics of features extracted from baseline periods \cite{pati2020quantitative, bogaarts2016improved, bai2016normalization, shedeed2016brain}. The main goal of this strategy is to emphasize changes in the features that correspond to factors that were modulated during the experimental task. In the case of the WAUC dataset, normalizing the features with respect to the first baseline period (baseline 1) highlights changes on the PSD due to both mental and physical stimuli. In turn, normalization with respect to the statistics of recording collected during the second baseline highlights modifications stemming only from mental workload changes, as only physical strain was modulated during this step. 

While commonly believed to improve classification accuracy, it is not clear from a statistical learning perspective whether and why these different normalization strategies work. Here, we quantitatively assess the impact that normalization has on mental workload performance under the lens of conditional and marginal shifts, as well as of cross-subject classification performance. As such, we perform a subject-wise normalization of each feature according to,

\begin{equation}
x^\prime_n = \frac{x_n - \beta}{\gamma^2},     
\end{equation}

where $\beta$ corresponds to the average feature vector and $\gamma$ the standard deviation considering the data recorded for the respective subject during the baseline periods. 

In addition to the aforementioned normalization strategies, we also perform experiments with features obtained after per-subject whitening of the data i.e., $\beta$ is the sample average and $\gamma$ the standard deviation for a given subject. This procedure is commonly referred to as z-score normalization. Lastly, we considered features without any normalization. As such, a total of four feature spaces are considered across our experiments: no normalization, whitening, and baselines 1/2 normalization. 

\subsection{Cross-subject mental workload classification}
In addition to analyzing the estimated cross-subject conditional and marginal shift for a mental workload assessment task, we also evaluate the cross-subject classification performance in this scenario. For that, we consider a leave-one-subject-out (LOSO) cross-validation scheme and train a different classifier per subject not included in the training set. Using this approach, we set our problem as a single-source single-target domain adaptation, where the source domain corresponds to the data of the all subjects pooled together, and the target domain corresponding to the subject left out as the test set. Although this is the setting considered in the experiments, we did not apply any domain adaptation scheme when learning classifiers since our objective in this work is to investigate distributional shifts and their relationship with out-of-distribution generalization.  

\subsection{Implementation details}
We implemented all classifiers, normalization, and cross-validation schemes using Scikit-learn \cite{pedregosa2011scikit}. For all experiments, we performed 30 independent repetitions considering slightly different partitions of the available data examples by randomly selecting 300 data points out of the 600 total available per subject/session. To enforce reproducibility, the random seed for all experiments was set to 10. The code corresponding to the following experiments are available on GitHub\footnote{\url{https://github.com/belaalb/EEG-DA}}.

A Random Forest with 20 estimators is used as the subject classifier to estimate $d_{\mathcal{H}}$ for computing the marginal shift. For predicting mental workload using LOSO cross-validation, we also use a Random Forest classifier, but in this case with 30 estimators.    

\section{Results and Discussion} \label{sec:results}
In this Section, we aim at answering the following main questions: i) Do different feature normalization schemes yield different values of distributional shifts? ii) Can the estimation of distributional shifts indicate how difficult it is to learn BCIs that generalize well on unseen subjects? iii) For a fixed feature space, are our findings consistent across two partitions of the WAUC containing subjects that had physical activity levels modulated by either bike or treadmill?

\subsection{Statistical shifts estimation}
\begin{figure}
    \centering
    \includegraphics[width=0.95\columnwidth]{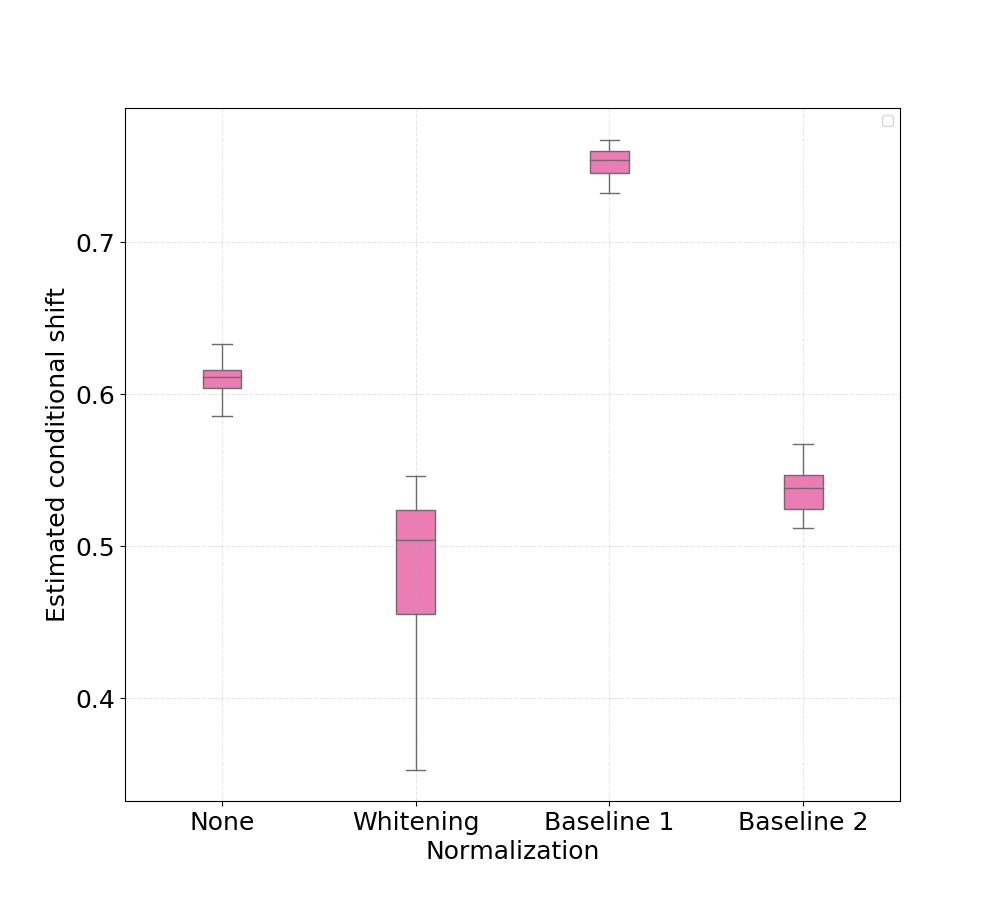}
    \caption{Boxplots with 30 independent estimates of the aggregate cross-subject conditional shift across different normalization strategies for participants which performed physical activity using a \textbf{treadmill}. Lower values represent smaller estimated conditional shift.}
    \label{fig:tread_cond_shift}
\end{figure}

\begin{figure}
    \centering
    \includegraphics[width=0.95\columnwidth]{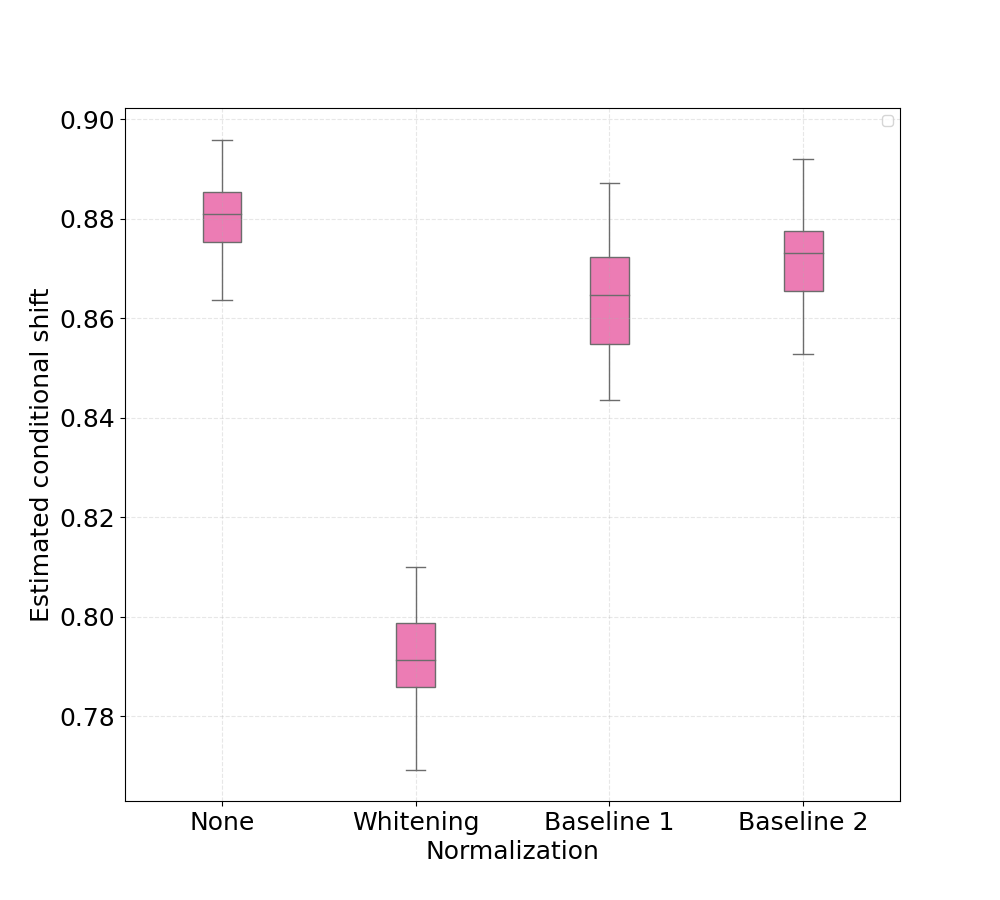}
    \caption{Boxplots with 30 independent estimates of the aggregate cross-subject conditional shift across different normalization strategies for participants which performed physical activity using a \textbf{bike}. Lower values represent smaller estimated conditional shift.}
    \label{fig:bike_cond_shift}
\end{figure}

Figures \ref{fig:tread_cond_shift} and \ref{fig:bike_cond_shift} show the boxplots with 30 estimates of the conditional shift for subjects corresponding to treadmill and bike, respectively. Considering the results obtained with the non-normalized version of the features as reference, it is possible to observe that whitening the features significantly improved the estimated aggregate conditional shift values (Eq. \ref{eq:frob}) for both treadmill and bike cases. As expected, this type of normalization is widely used in machine learning and known to improve overall classification performance in different applications of EEG data \cite{cruz2021self, sulaiman2010stress, zhang2013z}. 

In the case of normalizing the features with respect to the baseline periods, our findings show large differences when comparing the treadmill and bike conditions. For the bike case, normalizing the features yielded only a slight decrease in the observed conditional shift for both baseline 1 and 2 periods. For the treadmill condition, on the other hand, normalizing relative to baseline 1 (no physical activity) resulted in an increase of the aggregated conditional shift, thus potentially negatively affecting the performance of the mental workload assessment model to unseen subjects. Baseline 2 normalization, in turn, reduced the estimated conditional shift to levels closer to that achieved with per-subject whitening.

\begin{figure*}
	\centering
	\subfloat[tread_disp_no][No normalization.]{\includegraphics[width=0.47\textwidth]{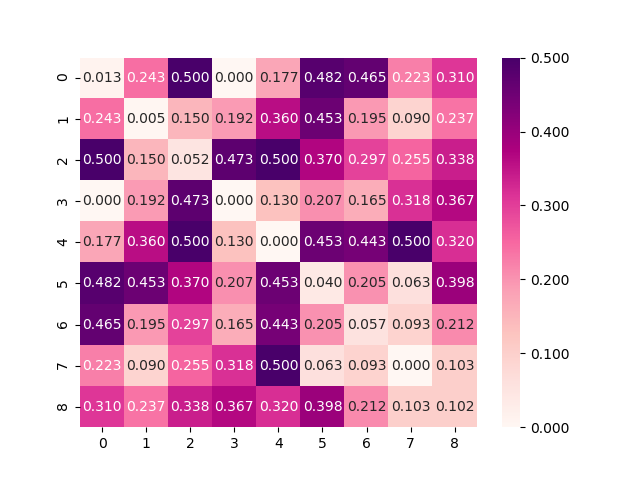}}
	\quad
	\subfloat[tread_disp_white][Whitening.]{\includegraphics[width=0.47\textwidth]{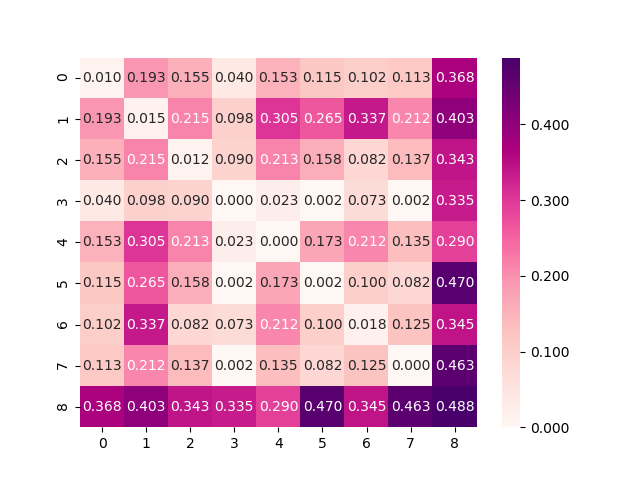}}
	\caption{Pair-wise cross-subject conditional shift with non-normalized and whitened features computed from subjects that performed physical activity on the \textbf{treadmill}.} 
	\label{fig:tread_disp_matrices}
\end{figure*}

\begin{figure*}
	\centering
	\subfloat[bike_disp_no][No normalization.]{\includegraphics[width=0.47\textwidth]{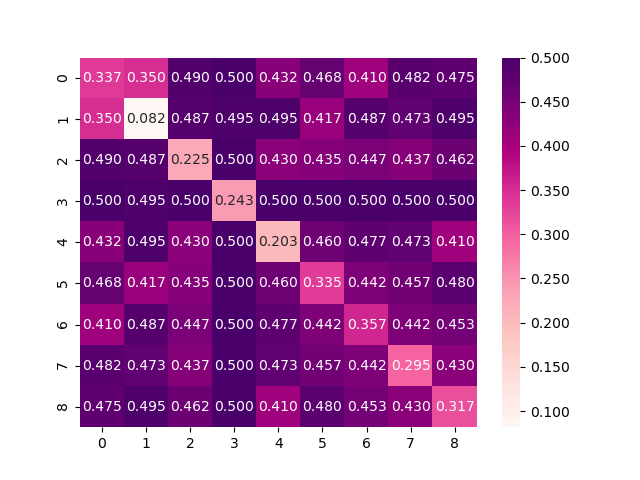}}
	\quad
	\subfloat[bike_disp_white][Whitening.]{\includegraphics[width=0.47\textwidth]{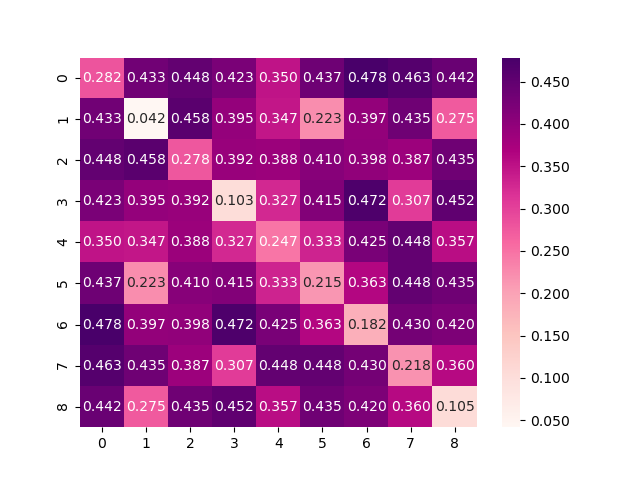}}
	\caption{Pair-wise cross-subject conditional shift with non-normalized and whitened features computed from subjects that performed physical activity on the \textbf{bike}.} 
	\label{fig:bike_disp_matrices}
\end{figure*}

In addition to investigating the aggregated conditional shift values, an in-depth analysis is also performed for the conditional shift values across all pairs of subjects in order to better understand the effects of feature normalization and the dependency on activity type. For that,  Figures~\ref{fig:tread_disp_matrices} and ~\ref{fig:bike_disp_matrices} display the disparity matrices $D$ computed considering features without normalization and whitening for both activity types, respectively. Notice that the entries at the main diagonal (i.e., within-subject disparity) were computed by having disjoint training and test sets, thus these values provide information about how good the employed labeling function approximation was. Also, these results correspond to a single estimate, thus do not show the variability of the reported quantities as it is the case in Figures \ref{fig:tread_cond_shift} and \ref{fig:bike_cond_shift}. 

It can be observed that the cross-subject conditional shift for the bike condition is much higher in comparison to the treadmill condition. This observation agrees with the findings of \cite{albuquerque2020wauc} and \cite{ladouce2019mobile}, which observed that different methods for inducing physical activity generate different EEG responses. Our results indicate that in the case of PSD features, this difference can be observed in practice by EEG responses which are more subject-specific, resulting in lower classification performance for the case of performing activity with a stationary bike, as reported in \cite{albuquerque2020wauc}.

\begin{figure}
    \centering
    \includegraphics[width=0.95\columnwidth]{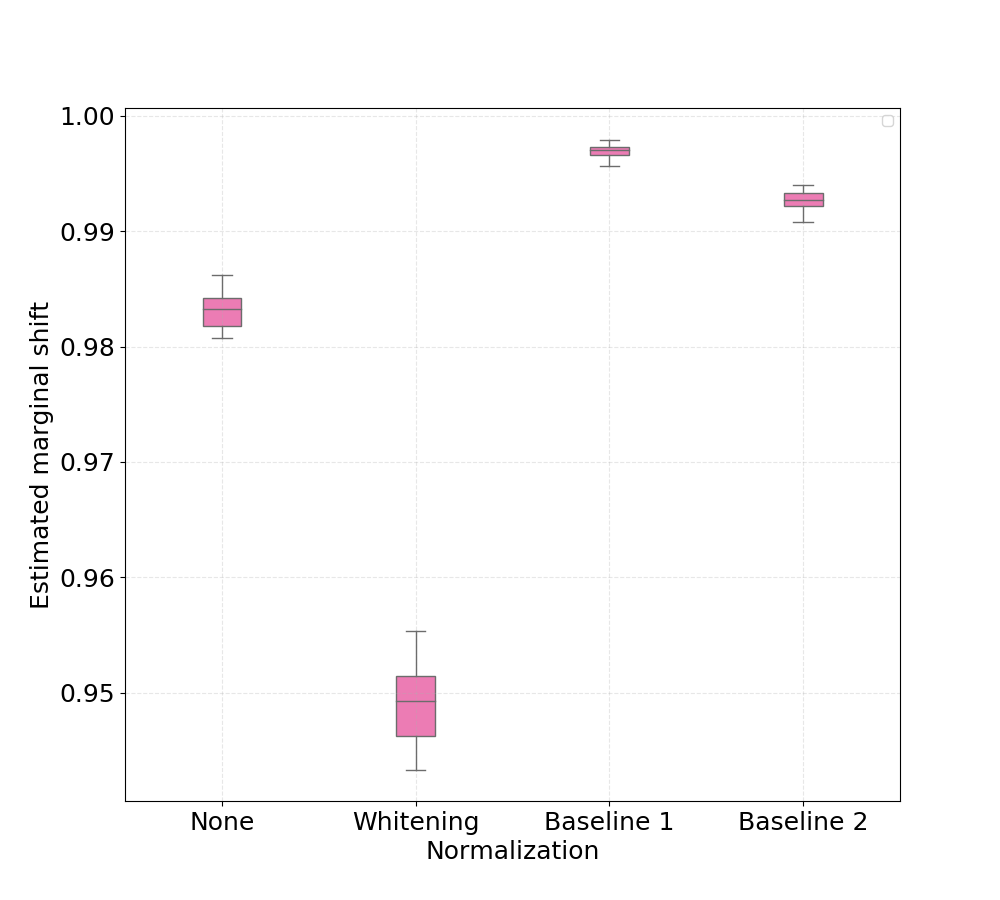}
    \caption{Boxplots with 30 independent estimates of the aggregate cross-subject marginal shift across different normalization strategies for participants which performed physical activity using a \textbf{treadmill}. Lower values represent smaller estimated marginal shifts.}
    \label{fig:tread_cov_shift}
\end{figure}
\begin{figure}
    \centering
    \includegraphics[width=0.95\columnwidth]{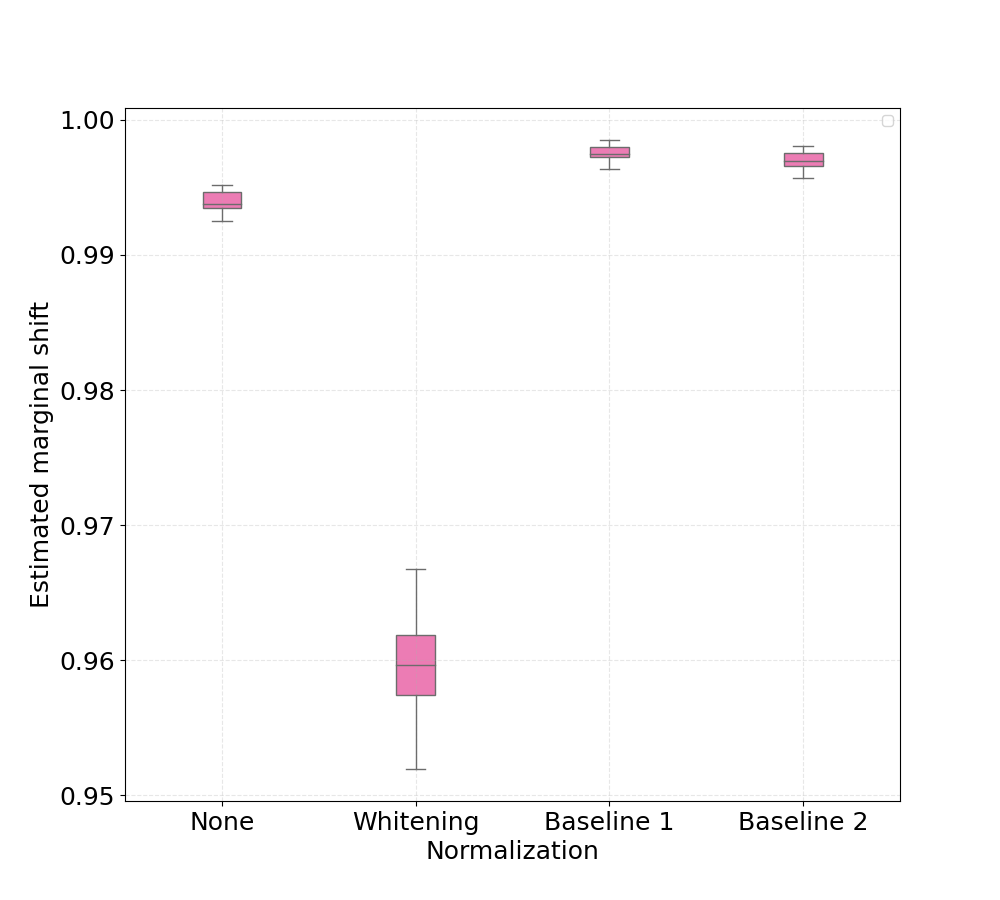}
    \caption{Boxplots with 30 independent estimates of the aggregate cross-subject marginal shift across different normalization strategies for participants which performed physical activity using a \textbf{bike}. Lower values represent smaller estimated marginal shifts.}
    \label{fig:bike_cov_shift}
\end{figure}

Similarly to the conditional shift analysis, we show in Figures \ref{fig:tread_cov_shift} and \ref{fig:bike_cov_shift} boxplots for the estimated aggregated marginal shift computed 30 times for all the considered normalization procedures, for treadmill and bike conditions, respectively. It is important to highlight that higher values of marginal shift (i.e., high $d_{\mathcal{H}}$) indicate a higher accuracy on pair-wise cross-subject classification. As such, discriminating data from two subjects in the PSD feature space consists in an easier task, and this contributes to higher cross-subject variability. We observe that for both treadmill and bike cases, subject-wise feature whitening decreased the estimated marginal shift, while baseline 1 and 2 normalization increased it. Intuitively, we expected z-score normalization to decrease the marginal shift, as the normalized features for all subjects have equal first and second order statistics. On the other hand, according to previous results on baseline normalization for EEG features, we expected that both baseline 1 and baseline 2 methods would make it more difficult for the classifier to discriminate subjects in the PSD feature space. 

\subsection{Generalization gap}
Lastly, target domain accuracy (i.e., test set or left-out subject) is reported for low/high mental workload classification using a LOSO cross-validation scheme. In addition to the test accuracy calculated on data from the subject left out, we also compute the classifier performance on the source domain by taking out from the training data 200 data points per subject. Based on the bound shown in Eq. \ref{eq:gordon}, our goal is to verify whether the estimated conditional and marginal shift values provide a way to assess the generalization gap between source and target domains. We use the training accuracy to compute the empirical risk, as it is equal to $1-\hat{R}_X[h_X]$ calculated with a 0-1 loss. Likewise, the true risk $R_X[h_X]$ was estimated as the accuracy on the test set. We calculated training and test average accuracy and the corresponding standard deviation across 30 independent runs. These values are shown per subject left out during training and averaged across all subjects. We also report average and standard deviation values of the generalization gap for each subject, calculated as the absolute difference between training and test accuracy. Tables~\ref{tab:tread_classif} and \ref{tab:bike_classif} present these quantities for the treadmill and the bike conditions, respectively.

\begin{table}
\centering
\caption{Results of binary mental workload classification with leave-one-subject-out cross validation for subjects that performed physical activity on the \textbf{treadmill}. For each subject, top and middle rows represent training and test accuracy, respectively. The estimated generalization gap is shown below the dotted line. Average and standard deviation across 30 independent runs are reported.}
\resizebox{0.99\columnwidth}{!}{
\begin{tabular}{cccccc}
\hline
  Subject                   & None              & Whitening         & Baseline 1        & Baseline 2        \\ \hline
\multirow{3}{*}{S0} & 0.974$\pm$0.004 & 0.936$\pm$0.007 & 0.985$\pm$0.003 & 0.982$\pm$0.004 \\
                    & 0.764$\pm$0.055 & 0.588$\pm$0.018 & 0.889$\pm$0.044 & 0.704$\pm$0.028 \\ \cdashline{2-5}\noalign{\vskip 0.4ex} 
                    & 0.210$\pm$0.055 & 0.348$\pm$0.018 & 0.096$\pm$0.044 & 0.279$\pm$0.029 \\ \hline
\multirow{3}{*}{S1} & 0.974$\pm$0.005 & 0.939$\pm$0.010 & 0.985$\pm$0.003 & 0.976$\pm$0.003 \\
                    & 0.543$\pm$0.043 & 0.628$\pm$0.042 & 0.550$\pm$0.037 & 0.560$\pm$0.051 \\ \cdashline{2-5}\noalign{\vskip 0.4ex}
                    & 0.431$\pm$0.045 & 0.311$\pm$0.044 & 0.435$\pm$0.037 & 0.416$\pm$0.050 \\ \hline
\multirow{3}{*}{S2} & 0.974$\pm$0.004 & 0.941$\pm$0.007 & 0.985$\pm$0.003 & 0.979$\pm$0.005 \\
                    & 0.575$\pm$0.046 & 0.602$\pm$0.058 & 0.524$\pm$0.015 & 0.630$\pm$0.052 \\ \cdashline{2-5}\noalign{\vskip 0.4ex}
                    & 0.399$\pm$0.046 & 0.340$\pm$0.060 & 0.461$\pm$0.016 & 0.349$\pm$0.051 \\ \hline
\multirow{3}{*}{S3} & 0.974$\pm$0.005 & 0.934$\pm$0.008 & 0.984$\pm$0.003 & 0.978$\pm$0.004 \\
                    & 0.700$\pm$0.079 & 0.968$\pm$0.060 & 0.643$\pm$0.082 & 0.603$\pm$0.055 \\ \cdashline{2-5}\noalign{\vskip 0.4ex}
                    & 0.249$\pm$0.063 & 0.054$\pm$0.065 & 0.292$\pm$0.104 & 0.281$\pm$0.093 \\ \hline
\multirow{3}{*}{S4} & 0.977$\pm$0.003 & 0.939$\pm$0.008 & 0.985$\pm$0.003 & 0.983$\pm$0.004 \\
                    & 0.662$\pm$0.032 & 0.771$\pm$0.056 & 0.540$\pm$0.022 & 0.541$\pm$0.024 \\ \cdashline{2-5}\noalign{\vskip 0.4ex} 
                    & 0.315$\pm$0.032 & 0.168$\pm$0.056 & 0.445$\pm$0.022 & 0.441$\pm$0.024 \\ \hline
\multirow{3}{*}{S5} & 0.973$\pm$0.003 & 0.942$\pm$0.009 & 0.989$\pm$0.004 & 0.979$\pm$0.004 \\
                    & 0.601$\pm$0.044 & 0.851$\pm$0.067 & 0.530$\pm$0.030 & 0.554$\pm$0.030 \\ \cdashline{2-5}\noalign{\vskip 0.4ex} 
                    & 0.372$\pm$0.044 & 0.092$\pm$0.067 & 0.454$\pm$0.030 & 0.425$\pm$0.028 \\ \hline
\multirow{3}{*}{S6} & 0.980$\pm$0.005 & 0.945$\pm$0.009 & 0.987$\pm$0.003 & 0.978$\pm$0.005 \\
                    & 0.751$\pm$0.042 & 0.595$\pm$0.037 & 0.588$\pm$0.074 & 0.567$\pm$0.049 \\ \cdashline{2-5}\noalign{\vskip 0.4ex} 
                    & 0.229$\pm$0.043 & 0.350$\pm$0.039 & 0.399$\pm$0.074 & 0.411$\pm$0.049 \\ \hline
\multirow{3}{*}{S7} & 0.973$\pm$0.004 & 0.935$\pm$0.007 & 0.985$\pm$0.003 & 0.975$\pm$0.004 \\
                    & 0.613$\pm$0.088 & 0.862$\pm$0.044 & 0.821$\pm$0.074 & 0.565$\pm$0.047 \\ \cdashline{2-5}\noalign{\vskip 0.4ex}
                    & 0.360$\pm$0.089 & 0.073$\pm$0.047 & 0.164$\pm$0.075 & 0.410$\pm$0.047 \\ \hline
\multirow{3}{*}{S8} & 0.984$\pm$0.003 & 0.959$\pm$0.006 & 0.989$\pm$0.003 & 0.982$\pm$0.003 \\ 
                    & 0.608$\pm$0.058 & 0.508$\pm$0.004 & 0.597$\pm$0.054 & 0.584$\pm$0.061 \\ \cdashline{2-5}\noalign{\vskip 0.4ex} 
                    & 0.375$\pm$0.059 & 0.451$\pm$0.006 & 0.392$\pm$0.055 & 0.398$\pm$0.060 \\ \hline
\multirow{3}{*}{All}& 0.976$\pm$0.005 & 0.941$\pm$0.011 & 0.985$\pm$0.004 & 0.979$\pm$0.005 \\
                    & 0.649$\pm$0.093 & 0.708$\pm$0.155 & 0.637$\pm$0.150 & 0.600$\pm$0.078 \\ \cdashline{2-5}\noalign{\vskip 0.4ex}
                    & 0.327$\pm$0.093 & 0.242$\pm$0.147 & 0.348$\pm$0.140 & 0.379$\pm$0.078 \\ \hline \hline
Cond. shift         & 0.608$\pm$0.013 & 0.482$\pm$0.060 & 0.753$\pm$0.009 & 0.537$\pm$0.014 \\
Marg. shift.        & 0.981$\pm$0.002 & 0.949$\pm$0.003 & 0.997$\pm$0.001 & 0.993$\pm$0.001 \\ \hline 
\end{tabular}}
\label{tab:tread_classif}
\end{table}

\begin{table}
\caption{Results of binary mental workload classification with leave-one-subject-out cross validation for subjects that performed physical activity on the \textbf{bike}. For each subject, top and middle rows represent training and test accuracy, respectively. The estimated generalization gap is shown below the dotted line. Average and standard deviation across 30 independent runs are reported.}
\resizebox{0.99\columnwidth}{5.21cm}{
\begin{tabular}{cccccc}
\hline
                         Subject                   & None              & Whitening         & Baseline 1        & Baseline 2        \\ \hline
\multirow{3}{*}{S0} & 0.921 $\pm$ 0.008 & 0.845 $\pm$ 0.009 & 0.923 $\pm$ 0.007 & 0.920 $\pm$ 0.006 \\
                    & 0.534 $\pm$ 0.026 & 0.536 $\pm$ 0.023 & 0.525 $\pm$ 0.016 & 0.558 $\pm$ 0.022 \\ \cdashline{2-5}\noalign{\vskip 0.4ex}
                    & 0.388 $\pm$ 0.028 & 0.309 $\pm$ 0.026 & 0.398 $\pm$ 0.016 & 0.363 $\pm$ 0.024 \\ \hline
\multirow{3}{*}{S1} & 0.893 $\pm$ 0.009 & 0.826 $\pm$ 0.015 & 0.899 $\pm$ 0.008 & 0.892 $\pm$ 0.007 \\ 
                    & 0.545 $\pm$ 0.041 & 0.579 $\pm$ 0.027 & 0.550 $\pm$ 0.046 & 0.568 $\pm$ 0.055 \\ \cdashline{2-5}\noalign{\vskip 0.4ex}
                    & 0.348 $\pm$ 0.043 & 0.246 $\pm$ 0.030 & 0.348 $\pm$ 0.047 & 0.324 $\pm$ 0.053 \\ \hline
\multirow{3}{*}{S2} & 0.906 $\pm$ 0.007 & 0.829 $\pm$ 0.011 & 0.909 $\pm$ 0.008 & 0.904 $\pm$ 0.009 \\
                    & 0.545 $\pm$ 0.037 & 0.550 $\pm$ 0.026 & 0.507 $\pm$ 0.007 & 0.519 $\pm$ 0.014 \\ \cdashline{2-5}\noalign{\vskip 0.4ex}
                    & 0.361 $\pm$ 0.038 & 0.279 $\pm$ 0.025 & 0.402 $\pm$ 0.012 & 0.385 $\pm$ 0.018 \\ \hline
\multirow{3}{*}{S3} & 0.892 $\pm$ 0.009 & 0.814 $\pm$ 0.012 & 0.896 $\pm$ 0.009 & 0.895 $\pm$ 0.009 \\ 
                    & 0.541 $\pm$ 0.039 & 0.681 $\pm$ 0.067 & 0.613 $\pm$ 0.078 & 0.578 $\pm$ 0.063 \\ \cdashline{2-5}\noalign{\vskip 0.4ex}
                    & 0.351 $\pm$ 0.038 & 0.133 $\pm$ 0.067 & 0.284 $\pm$ 0.079 & 0.317 $\pm$ 0.063 \\ \hline
\multirow{3}{*}{S4} & 0.903 $\pm$ 0.008 & 0.836 $\pm$ 0.013 & 0.907 $\pm$ 0.008 & 0.900 $\pm$ 0.007 \\ 
                    & 0.549 $\pm$ 0.027 & 0.541 $\pm$ 0.024 & 0.575 $\pm$ 0.054 & 0.542 $\pm$ 0.034 \\ \cdashline{2-5}\noalign{\vskip 0.4ex}
                    & 0.354 $\pm$ 0.029 & 0.295 $\pm$ 0.028 & 0.331 $\pm$ 0.051 & 0.358 $\pm$ 0.036 \\ \hline
\multirow{3}{*}{S5} & 0.910 $\pm$ 0.009 & 0.837 $\pm$ 0.012 & 0.914 $\pm$ 0.007 & 0.909 $\pm$ 0.008 \\ 
                    & 0.529 $\pm$ 0.020 & 0.555 $\pm$ 0.037 & 0.531 $\pm$ 0.019 & 0.522 $\pm$ 0.019 \\ \cdashline{2-5}\noalign{\vskip 0.4ex}
                    & 0.380 $\pm$ 0.023 & 0.283 $\pm$ 0.040 & 0.383 $\pm$ 0.021 & 0.387 $\pm$ 0.020 \\ \hline
\multirow{3}{*}{S6} & 0.914 $\pm$ 0.008 & 0.847 $\pm$ 0.013 & 0.918 $\pm$ 0.008 & 0.914 $\pm$ 0.007 \\ 
                    & 0.529 $\pm$ 0.022 & 0.520 $\pm$ 0.015 & 0.535 $\pm$ 0.026 & 0.536 $\pm$ 0.025 \\ \cdashline{2-5}\noalign{\vskip 0.4ex}
                    & 0.385 $\pm$ 0.022 & 0.327 $\pm$ 0.020 & 0.383 $\pm$ 0.027 & 0.378 $\pm$ 0.027 \\ \hline
\multirow{3}{*}{S7} & 0.900 $\pm$ 0.007 & 0.841 $\pm$ 0.009 & 0.905 $\pm$ 0.007 & 0.898 $\pm$ 0.010 \\
                    & 0.549 $\pm$ 0.033 & 0.547 $\pm$ 0.026 & 0.553 $\pm$ 0.033 & 0.557 $\pm$ 0.043 \\ \cdashline{2-5}\noalign{\vskip 0.4ex}
                    & 0.350 $\pm$ 0.032 & 0.294 $\pm$ 0.027 & 0.352 $\pm$ 0.033 & 0.341 $\pm$ 0.043 \\ \hline
\multirow{3}{*}{S8} & 0.896 $\pm$ 0.009 & 0.841 $\pm$ 0.012 & 0.904 $\pm$ 0.008 & 0.900 $\pm$ 0.008 \\
                    & 0.551 $\pm$ 0.039 & 0.599 $\pm$ 0.030 & 0.546 $\pm$ 0.033 & 0.542 $\pm$ 0.028 \\ \cdashline{2-5}\noalign{\vskip 0.4ex}
                    & 0.345 $\pm$ 0.040 & 0.242 $\pm$ 0.034 & 0.358 $\pm$ 0.033 & 0.358 $\pm$ 0.031 \\ \hline
\multirow{3}{*}{All} & 0.904 $\pm$ 0.012 & 0.835 $\pm$ 0.016 & 0.908 $\pm$ 0.011 & 0.904 $\pm$ 0.012 \\
                     & 0.541 $\pm$ 0.033 & 0.567 $\pm$ 0.057 & 0.548 $\pm$ 0.050 & 0.547 $\pm$ 0.042 \\ \cdashline{2-5}\noalign{\vskip 0.4ex}
                     & 0.363 $\pm$ 0.037 & 0.268 $\pm$ 0.065 & 0.360 $\pm$ 0.054 & 0.357 $\pm$ 0.045 \\ \hline \hline
Cond. shift          & 0.880 $\pm$ 0.008 & 0.792 $\pm$ 0.010 & 0.864 $\pm$ 0.011 & 0.872 $\pm$ 0.010 \\
Marg. shift.         & 0.994 $\pm$ 0.001 & 0.959 $\pm$ 0.004 & 0.998 $\pm$ 0.001 & 0.997 $\pm$ 0.001 \\ \hline                      
\end{tabular}}
\label{tab:bike_classif}
\end{table}

According to the results presented in Table \ref{tab:tread_classif}, we observe that, as predicted by the bound in Eq. \ref{eq:gordon}, z-score normalization, i.e., the features with lower conditional and marginal shifts, presented the smallest approximated generalization gap between source and target domains. This finding is similarly observed in the case of the group of subjects that performed the experiment with the stationary bike, as shown by the results reported in Table \ref{tab:bike_classif}. An overall comparison between treadmill and bike subjects also reveals that inter-subject generalization, as measured by the estimate of the risk on the source domain (training subjects), is considerably lower for the bike condition. This aspect could also have been predicted by the diagonal values of the disparity matrix (Figures  \ref{fig:tread_disp_matrices} and \ref{fig:bike_disp_matrices}) which show that for the majority of the subjects the approximated labeling function seems to be easier to approximate for the treadmill condition.  

Moreover, in the case of the treadmill group, we observe that baseline 1 normalization yielded a slightly smaller average generalization gap in comparison to baseline 2, even though it presented a considerably higher conditional shift. As both normalization strategies obtained close values of average marginal shift, we believe this indicates that the two analyzed statistical shifts might differ in their contribution to the generalization bound. Furthermore, considering the average results across all subjects, z-score normalization presented the best performance in terms of accuracy, being able to correctly classify roughly 70\% of points from subjects not considered during training. It is important to highlight that as opposed to normalizing with respect to baseline recordings, which requires a calibration step to collect data prior to the actual task, z-score normalization does not need any extra information other than the features extracted from data corresponding to the task. On the other hand, despite better mitigating cross-subject variability and being more efficient in terms of data collection time, the intra-subject classification performance of models trained on z-score normalized features is worse in comparison with other strategies, indicating there might be a trade-off between improving cross-subject performance and maintaining good accuracy on the source domains.

\begin{figure}
    \centering
    \includegraphics[width=0.95\columnwidth]{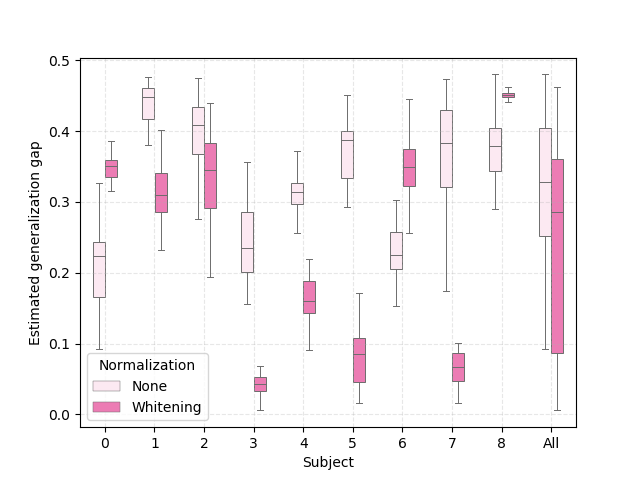}
    \caption{Boxplot with 30 independent estimates of the generalization gap for the subjects that performed the experiment using a \textbf{treadmill}. The generalization gap is computed as the difference between training and test accuracy using a leave-one-subject-out cross-validation setting.}
    \label{fig:tread_gen_gap}
\end{figure}

\begin{figure}
    \centering
    \includegraphics[width=0.95\columnwidth, height=4.5cm]{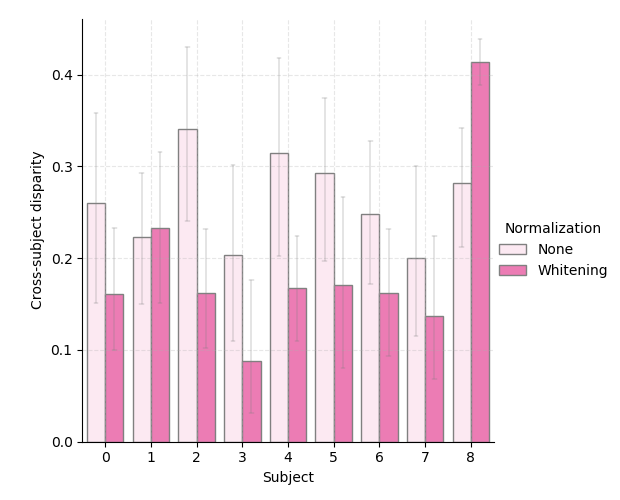}
    \caption{Bar plot with the average cross-subject disparity for 30 independent estimates of the disparity matrix for the subjects that performed the experiment using a \textbf{treadmill}.}
    \label{fig:tread_cond_shift_persub}
\end{figure}

To provide further empirical evidence that the analysis of the statistical shifts as employed in this work can be used to select a feature normalization that yields better cross-domain (i.e., cross-subject) generalization, we show in Fig. \ref{fig:tread_gen_gap} boxplots of 30 independent generalization gap estimates for each subject within the treadmill group. 
In addition, we provide in Fig.~\ref{fig:tread_cond_shift_persub} a bar plot with average values of cross-subject disparity for all subjects that had physical workload modulated by the treadmill. These values were computed using the columns of the average disparity matrix resulting from the 30 repetitions executed to generate Fig.~\ref{fig:tread_cond_shift}. Notice that within this analysis we are not taking into account the marginal shift. By comparing Figs.~\ref{fig:tread_gen_gap} and ~\ref{fig:tread_cond_shift_persub} we observe that for subjects 2, 3, 4, 5, 7, and 8 the normalization method with lower average conditional shift, yielded a smaller median estimated generalization gap. Importantly, we observe that subject 8 did not benefit from z-score normalization, as the conditional shift increased, along with an increase in the generalization and a decrease in the accuracy as shown in Table ~\ref{tab:tread_classif}.


\subsection{Main takeaways} 
In light of our results and discussion, we highlight the observations we found most relevant to be considered by future research. In case the goal is to improve out-of-distribution performance, normalization procedures that decrease the overall cross-subject conditional shift should be prioritized since they yield smaller generalization gaps. For devising passive BCIs with the aim of monitoring mental workload under physical activity, our analysis showed that z-score normalization provided the best strategy for normalizing EEG power spectral density features. Moreover, such normalized feature spaces should be considered in case representation learning based on domain adaptation are used to learn domain-invariant classifiers. Notice there is a caveat that should also be taken into account: the results shown in Tables \ref{tab:tread_classif} and \ref{tab:bike_classif} consistently indicate (i.e. across equipment for modulating physical activity and normalization procedures) that improving out-of-distribution performance via normalizing the features leads to a decrease on the model accuracy computed on unseen data from the training subjects.  

\section{Conclusions}\label{sec:conc}

In this work, we present the first steps towards better understanding the cross-subject variability phenomena seen with passive EEG-based BCIs from a statistical learning perspective. We looked at this problem through the lens of domain adaptation and proposed strategies to estimate distributional shifts between conditional and marginal distributions corresponding to the data generating process of features and labels from different subjects. To evaluate the proposed approach, the WAUC dataset was used and binary mental workload assessment from EEG power spectral features was performed. Our analysis showed that feature normalization, as well as data collection conditions such as the equipment used to induce physical workload, had a relevant impact in the estimated values of conditional shift. Importantly, our results showed that whitening the features (i.e., performing z-score normalization) mitigated both conditional and marginal shifts and improved mental workload assessment on unseen subjects at training time. Future work consists on employing the developed strategies to estimate distributional shifts in order to better inform the development of domain adaptation methods for EEG applications.


\bibliographystyle{IEEEtran}
\bibliography{bibliography}

\end{document}